\documentclass[letterpaper]{article} 
\usepackage{aaai25}  
\usepackage{times}  
\usepackage{helvet}  
\usepackage{courier}  
\usepackage[hyphens]{url}  
\usepackage{graphicx} 
\usepackage{natbib}  
\usepackage{caption} 
\usepackage{algorithm}
\usepackage{algorithmic}

\usepackage{newfloat}
\usepackage{listings}
\DeclareCaptionStyle{ruled}{labelfont=normalfont,labelsep=colon,strut=off} 
\floatstyle{ruled}
\newfloat{listing}{tb}{lst}{}
\floatname{listing}{Listing}
\usepackage{colortbl}
\definecolor{myy}{RGB}{126,95,0}
\definecolor{mygray}{gray}{.9}
\definecolor{bblue}{RGB}{30,80,120}
\definecolor{mygray1}{gray}{.7}
\usepackage{bm}
\usepackage{mathtools}
\usepackage{subcaption}
\usepackage[nopar]{lipsum}
\usepackage[export]{adjustbox}
\usepackage{bbold}
\usepackage{utfsym}
\usepackage{url}
\usepackage{multirow}
\usepackage{pifont}

\definecolor{mygray}{RGB}{127,127,127}
\definecolor{mygreen}{RGB}{93,174,86}

\makeatletter
\newcommand{\thickhline}{%
	\noalign {\ifnum 0=`}\fi \hrule height 1pt
	\futurelet \reserved@a \@xhline
}
\makeatother

\usepackage[capitalize]{cleveref}
\crefname{section}{Sec.}{Secs.}
\Crefname{section}{Section}{Sections}
\Crefname{table}{Table}{Tables}
\crefname{table}{Tab.}{Tabs.}

\RequirePackage{xspace}
\makeatletter
\DeclareRobustCommand\onedot{\futurelet\@let@token\@onedot}
\def\@onedot{\ifx\@let@token.\else.\null\fi\xspace}
\def\eg{\emph{e.g}\onedot} 

\def\ie{\emph{i.e}\onedot}

\makeatother

\title{Language Prompt  for Autonomous Driving}
\author{
    Dongming Wu\textsuperscript{\rm 1}\equalcontrib \footnote{The work is done during Dongming Wu interned at MEGVII.}, 
    Wencheng Han\textsuperscript{\rm 2}\equalcontrib, 
    Yingfei Liu\textsuperscript{\rm 3}, \\ 
    Tiancai Wang\textsuperscript{\rm 3}, 
    Cheng-zhong Xu\textsuperscript{\rm 2}, 
    Xiangyu Zhang\textsuperscript{\rm 3,}\textsuperscript{\rm 4}, 
    Jianbing Shen\textsuperscript{\rm 2}\thanks{Corresponding author: \textit{Jianbing Shen}.}
}

\affiliations{
    
    \textsuperscript{\rm 1} Beijing Institute of Technology, \\
    \textsuperscript{\rm 2} SKL-IOTSC, CIS, University of Macau, \\
    \textsuperscript{\rm 3} MEGVII Technology, \\
    \textsuperscript{\rm 4} Beijing Academy of Artificial Intelligence \\

    {\{wudongming97, wencheng256\}@gmail.com,}
    {wangtiancai@megvii.com,} {jianbingshen@um.edu.mo}
}

\usepackage{bibentry}

\begin{document}

\maketitle

\begin{abstract}
A new trend in the computer vision community is to capture objects of interest following flexible human command represented by a natural language prompt. However, the progress of using language prompts in driving scenarios is stuck in a bottleneck due to the scarcity of paired prompt-instance data. To address this challenge, we propose the first object-centric language prompt set for driving scenes within  3D, multi-view, and multi-frame space, named NuPrompt. It expands nuScenes dataset by constructing a total of 40,147 language descriptions, each referring to an average of 7.4 object tracklets. Based on the object-text pairs from the new benchmark, we formulate a novel prompt-based driving task, \ie, employing a language prompt to predict the described object trajectory across views and frames. Furthermore, we provide a simple end-to-end baseline model based on Transformer, named PromptTrack. Experiments show that our PromptTrack achieves impressive performance on NuPrompt. We hope this work can provide some new insights for the self-driving community.  The data and code have been released at https://github.com/wudongming97/Prompt4Driving.
\end{abstract}

\begin{figure*}[h]
\centering
\hsize=\textwidth
    \includegraphics[scale=0.85]{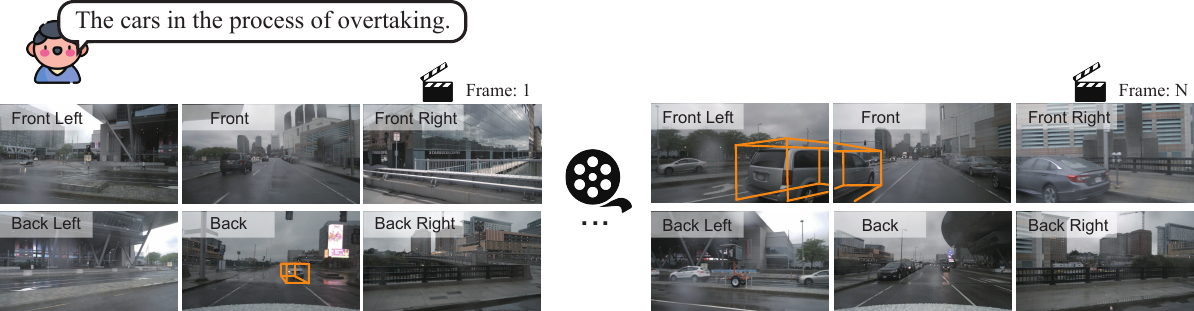}
    \caption{{A representative example from NuPrompt}. The language prompt ``the cars in the process of overtaking'' is precisely annotated and corresponded to the objects within the 3D, multi-frame, and multi-view driving space.    NuPrompt contains 40,147 object-prompt pairs with fine-grained semantic alignment. Each language prompt refers to multiple object trajectories.}
    \label{fig:examples}
\end{figure*}

\begin{table*}[t]
\centering
\small
\resizebox{\textwidth}{!}{
	\begin{tabular}{l||ccc|ccc|c}
	\hline \thickhline
	  \cellcolor[gray]{0.9} & \cellcolor[gray]{0.9} &\cellcolor[gray]{0.9} &\cellcolor[gray]{0.9} & \cellcolor[gray]{0.9} &\cellcolor[gray]{0.9} &\cellcolor[gray]{0.9} & \cellcolor[gray]{0.9}\\
    \rowcolor[gray]{0.9}
	\multirow{-2}{*}{Dataset} & \multirow{-2}{*}{Basic Task} &  \multirow{-2}{*}{3D} &   \multirow{-2}{*}{\#Views}   & \multirow{-2}{*}{\#Videos}   & \multirow{-2}{*}{\shortstack{\#Frames}}  & \multirow{-2}{*}{\shortstack{\#Prompts}}  & \multirow{-2}{*}{\shortstack{\# Instances\\per-prompt} }  \\
	\hline
       RefCOCO~\cite{yu2016modeling}  & Det\&Seg & \usym{2717} & 1  & -  & 26,711     & 142,209 & 1     \\
       Talk2Car~\cite{deruyttere2019talk2car} & Det & \usym{2713} & 1 &  -   & 9,217       &  11,959 & 1  \\ 
       Cityscapes-Ref~\cite{vasudevan2018object} & Det & \usym{2717} & 1 & - & 4,818 & 30,000 & 1    \\ 
        Refer-KITTI~\cite{wu2023referring}  & MOT & \usym{2717} & 1   & 18 & 6,650    &  818 & 10.7 \\ 
        NuScenes-QA~\cite{qian2023nuscenes} & VQA &  \usym{2713} & 6 & 850 &  34,149  & 459,941 & - \\
        DriveLM-nuScenes~\cite{drivelm} & Det\&VQA &  \usym{2717}  & 6  & - & 4,871 & 445,209 & -  \\
	\hline
 NuPrompt (Ours) & MOT &  \usym{2713} & 6 & 850 &  34,149 & 40,776  & 7.4  \\
 \hline  \thickhline
\end{tabular} }
 \caption{{Comparison of our NuPrompt with existing prompt-based datasets.} `-' means unavailable.  NuPrompt provides the nature and complexity of driving scenes, \ie, 3D, multi-view space, and multi-frame domain. Besides, it  focuses on object-centric understanding by pairing a language prompt with multiple targets of interest. 
}
\label{table:dataset}
\end{table*}

%

\section{Introduction}
\label{sec:intro}

Leveraging natural language description in visual tasks is one of the recent trends in the vision community~\cite{radford2021learning,kirillov2023segment}. 
It has garnered significant interest for its potential applications in various downstream tasks, such as embodied intelligence and human-robot interactions~\cite{deruyttere2019talk2car,chen2023end,gupta2023visual,hu2023planning,wu2023topomlp,wu20231st,bai2024}.
Its core idea is to predict the desired answer by shifting human instruction inputs but not updating model weights, delivering high adaptability in response to varying human demands.
A key factor for the progress in 2D scenes is the availability of large-scale image-text pairs~\cite{lin2014microsoft,changpinyo2021conceptual,schuhmann2021laion}.
However, this success is hard to replicate in self-driving scenarios due to the scarcity of 3D instance-text pairs.

Pioneering works like Talk2Car~\cite{deruyttere2019talk2car}, Cityscapes-Ref~\cite{vasudevan2018object} have started to incorporate natural language into object detection tasks in driving scenes. 
Unfortunately, these datasets only allow each expression to refer to a single object within an individual image, restricting their usage in scenarios with multiple referred objects or changing temporal states. 
Furthermore, Refer-KITTI~\cite{wu2023referring} addressed this issue by extending the KITTI dataset~\cite{geiger2012we} to include expressions that ground multiple video objects. This work mainly focuses on modular images and 2D detection, thereby leaving room for improvement in 3D driving scenes. A recent advancement, namely NuScenes-QA~\cite{qian2023nuscenes}, offers numerous question-answer pairs for 3D multi-view driving scenes, making significant strides in the use of language prompts. However, it primarily contributes to scene-level understanding and overlooks the direct and fine-grained semantic correspondence between 3D instances and natural language expression.

To advance the research of prompt learning in driving scenarios, we propose a new large-scale benchmark,  named \textbf{NuPrompt}. The benchmark is built on the popular multi-view 3D object detection dataset nuScenes~\cite{nuscenes2019}. 
We assign a language prompt to a collection of objects sharing the same characteristics for grounding them. 
Essentially, this benchmark provides lots of 3D instance-text pairings with three primary attributes:
\ding{182} \textit{Real-driving descriptions.} 
Different from existing benchmarks that only represent 2D objects from modular images, the prompts of our dataset characterize a diverse range of driving-related objects from 3D, panoramic views, and long-temporal space.
Fig.~\ref{fig:examples} shows a typical example, \ie,  a car surpasses our car from behind towards the front across multiple views.
\ding{183} \textit{Instance-level prompt annotations.}
Every prompt indicates a fine-grained and discriminative object-centric description, as well as enabling it to cover an arbitrary number of driving objects.
\ding{184} \textit{Large-scale language prompts.}
From a quantitative perspective, NuPrompt has 40,147 language prompts.

Along with NuPrompt, we present a new and challenging prompt-based driving perception and prediction task, whose main idea is to track and forecast the trajectory for 3D objects based on a specified language prompt.
The challenges of this task lie in two aspects: temporal association across frames and cross-modal semantic comprehension.
To address the challenges,  we propose an end-to-end baseline built on camera-only 3D tracker PF-Track~\cite{pang2023standing}, named \textbf{PromptTrack}.
Note that PF-Track has exhibited excellent spatial-temporal modeling through its past and future reasoning branches.
We additionally involve cross-attention between prompt embedding and visual features, and then add one prompt reasoning branch to ground prompt-referred objects.
Furthermore, we also evaluate the motion prediction of these prompt-referred objects in our experiments.
In addition to the prompt-based prediction tasks,  we expect that our annotations can facilitate future research in multi-model LLM of autonomous driving.

In summary, our contributions are three-fold: 

\begin{itemize}
  \item We propose a new large-scale language prompt set for driving scenes, named NuPrompt.  As far as we know, it is the first dataset specializing in multiple 3D objects of interest from video domain. 
  \item We construct a new prompt-based driving perception and prediction task, which requires using a language prompt as a semantic cue to predict object trajectories.
  \item 
  We develop a simple baseline model that unifies prompt-based object tracking and motion prediction in a single framework, called PromptTrack. 
\end{itemize}


 \begin{figure*}[t]
	\centering
	\resizebox{1\textwidth}{!}
	{
		\includegraphics[width = 16cm]{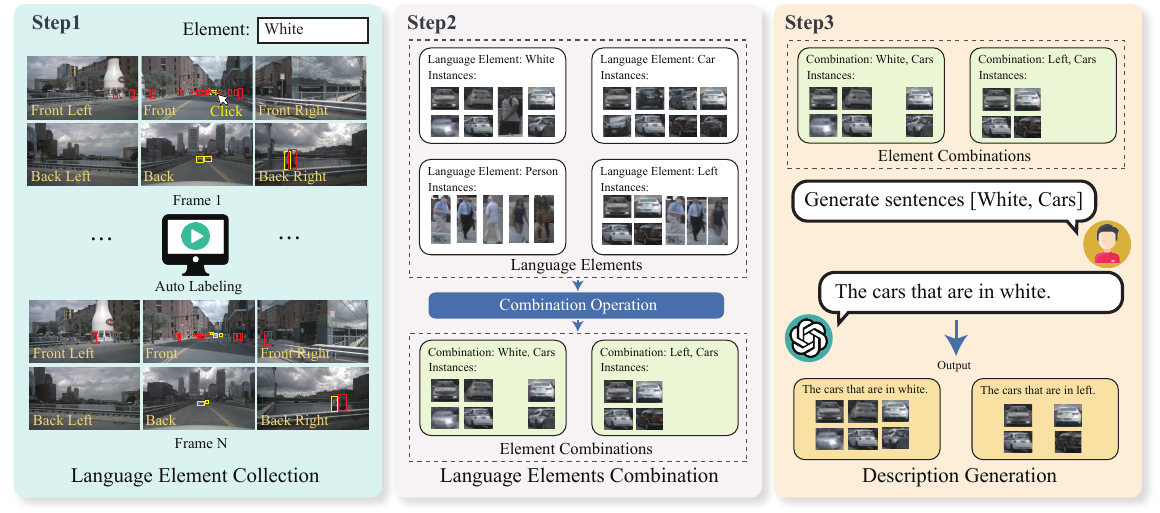}
	}
	\caption{{Pipeline of language prompt annotation procedure}, which includes three steps: language element collection, language element combination, and description generation. Firstly, we pair each language tag with referent objects during the language element collection phase. Following this, certain language elements are selected and combined in the language element combination stage. Finally, with the combinations obtained, we employ LLM to create language descriptions.
	}
	\label{fig:labeling_pipeline}
\end{figure*}

\section{Related Work}
\label{sec:relatedwork}

\noindent\textbf{Language Prompt in Driving Scenes.}
The utilization of human commands within driving scenes allows the system to understand driving systems from the human perspective, thereby facilitating human control over driving procedures. Talk2Car~\cite{deruyttere2019talk2car}, the pioneering benchmark featuring language prompts for autonomous vehicles, is constructed on the base of nuScenes~\cite{nuscenes2019}. However, its annotation only comprises keyframes that catch the eye of annotators.
However, the prompts deployed in both Talk2Car tend to represent an individual object.
To solve this problem,  Refer-KITTI~\cite{wu2023referring} further develops KITTI~\cite{geiger2012we}, where each prompt can refer to a collection of referent objects.  
More recently,  NuScenes-QA~\cite{qian2023nuscenes} opens a new avenue, Visual Question Answering (VQA), for understanding scene-level driving scenarios. 
DriveLM~\cite{drivelm} considers 2D crucial objects for graph visual question answering.
Although there are other natural language sets used in the driving area, like BDD-X~\cite{kim2018textual}, DRAMA~\cite{malla2023drama},  TalkBEV~\cite{talk2bev}, and Rank2Tell~\cite{rank2tell},  they are completely different from ours because of using language as captions for improving driving explanations.
A thorough comparison between existing prompt-based driving datasets and ours is summarized in Table~\ref{table:dataset}.

\label{Sec:collection}

\noindent\textbf{Referring Expression Understanding.}
Given a language prompt, the goal of referring expression understanding is to localize the described objects using boxes or masks, which shares a similar idea with our prompt-based driving benchmark. 
The initiation of datasets like RefCOCO/+/g~\cite{yu2016modeling} has helped stimulate interest in this field. These datasets use succinct yet unambiguous natural language expressions to ground a visual region within an image.
Some follow-up works further improve this dataset by allowing it to support expressions that refer to unlimited target objects~\cite{liu2023gres,chen2019weakly}. 
Besides, Refer-DAVIS$_{16/17}$~\cite{khoreva2019video} and Refer-Youtube-VOS~\cite{seo2020urvos} are another two popular video-based referring expression understanding benchmarks supporting video object segmentation.
A recent work in the field called GroOT~\cite{nguyen2023type} expands the large-scale multi-object tracking dataset TAO~\cite{dave2020tao} to support referring understanding.

\section{Dataset Overview}

\subsection{Data Collection and Annotation}
Our NuPrompt is built on one of the most popular datasets for multi-view 3D object detection, nuScenes~\cite{nuscenes2019}. While the original nuScenes dataset includes visual images and point cloud data, we here focus solely on visual images for NuPrompt. As shown in Fig.~\ref{fig:labeling_pipeline}, the cars collecting the data are equipped with six different cameras: Front, Front Left, Front Right, Back Right, Back Left, and Back. These cameras overlap in some areas. 
Therefore, NuPrompt provides 360$^{\circ}$ of 3D space for each scene.

To efficiently generate training labels for the new dataset, we designed a three-step semi-automatic labeling pipeline (see Fig.~\ref{fig:labeling_pipeline}).
The first step aims at identifying language elements and associating them with 3D bounding boxes. The second step is to combine the language elements using certain rules. In the third step, we base the language element combinations to produce various language prompts using a large language model (LLM). 
Detailed information about these three steps is provided as follows.

\begin{figure*}[t]
	\centering
   \resizebox{0.33\textwidth}{!}
	{
		\includegraphics[width = 16cm]{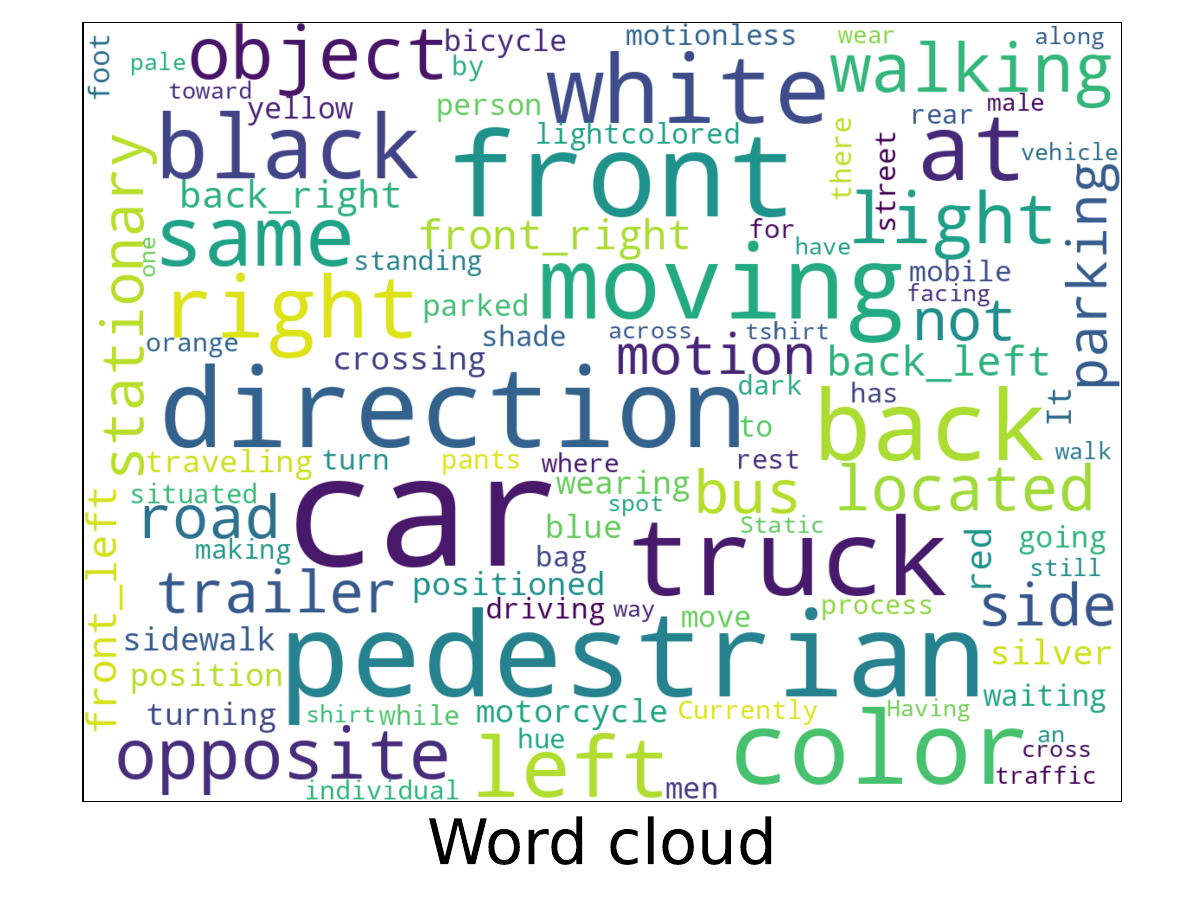}
	}
 	\resizebox{0.33\textwidth}{!}
	{
		\includegraphics[width = 16cm]{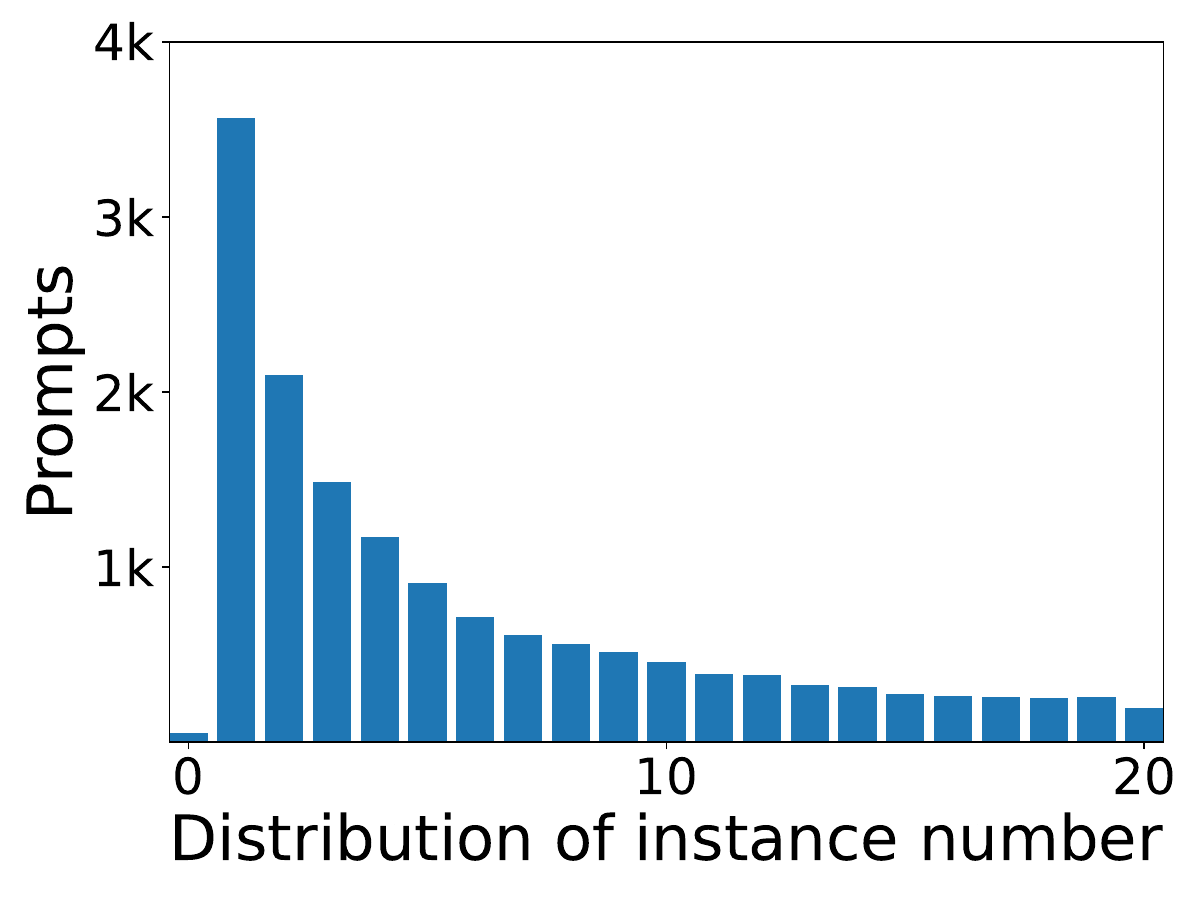}
	}
 	\resizebox{0.33\textwidth}{!}
	{
		\includegraphics[width = 16cm]{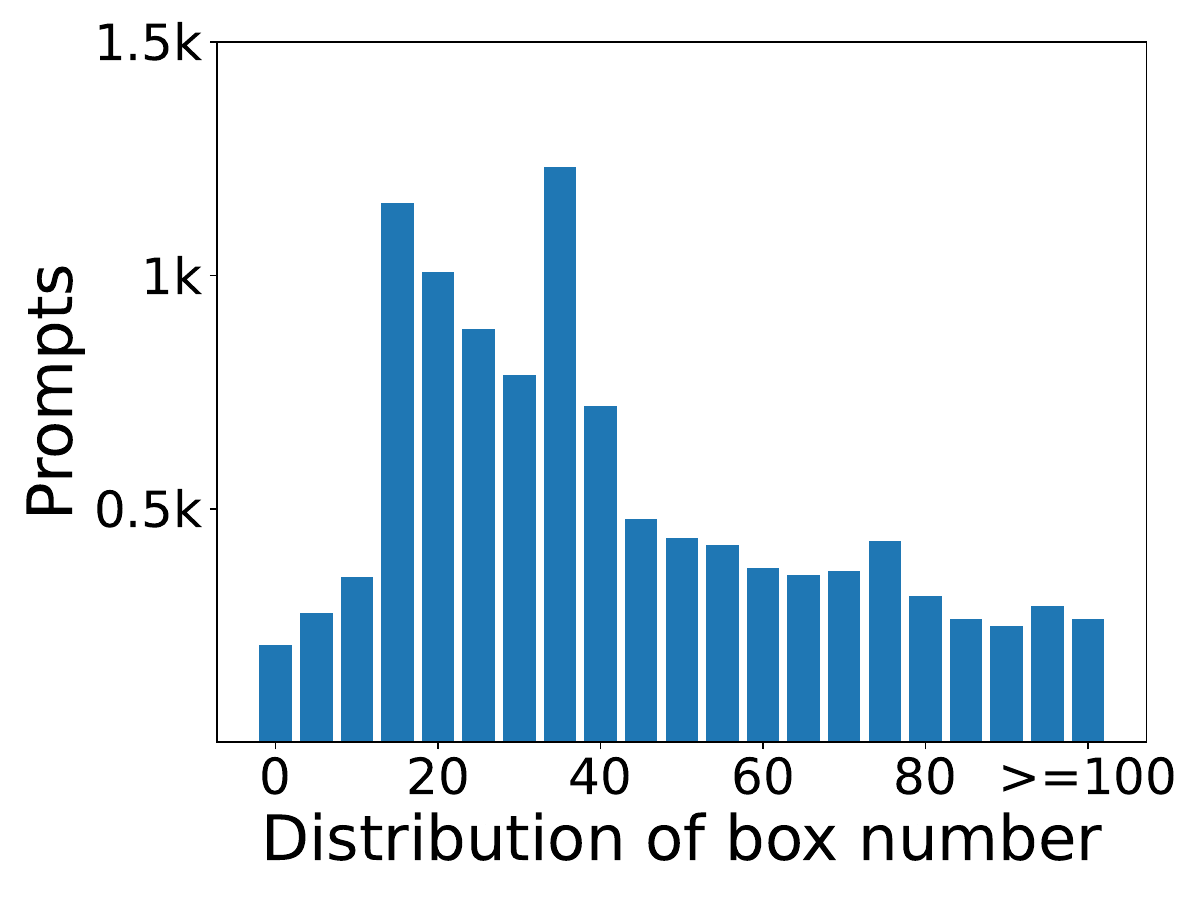}
	}
 	\caption{{Left}: Word cloud of top 100 words in NuPrompt. {Middle}: Distribution of instance number per prompt, where instance number also represents object tracklet number. {Right:}   Distribution of box number per prompt. 
	}
	\label{fig:dataset_summary}
\end{figure*} 

\noindent\textbf{Step 1: Language Element Collection.} 
This paper uses the term ``language element" to refer to a basic attribute of objects. Examples of language elements include colors (e.g. red, yellow, and black), actions (e.g. running, stopping, and crossing the road), locations (e.g. left, right, and back), and classes (e.g. car and pedestrian), which cover diverse descriptions of driving scenes.
The key problem is how to label the bounding boxes with the corresponding language elements.
To solve this, we design a labeling system to manually collect and match language elements with bounding boxes in a video sequence, as demonstrated in Fig.~\ref{fig:labeling_pipeline}. Annotators type the language element texts and click on the corresponding bounding boxes. When the target changes status and no longer belongs to the language elements, annotators need to click on the target again and remove it from the list.
This procedure can efficiently reduce the amount of human labor required. To ensure a variety of expressions, each video is assigned to five independent annotators who manually create descriptive expressions to formulate query sentences. Two other annotators then carefully check the matching between the expressions and the video targets.

\noindent\textbf{Step 2: Language Elements Combination.} As mentioned earlier, language elements are basic attributes of objects. By combining these attributes, we can create various descriptions for different groups of objects. There is one logical relationship we can use to merge the attributes: AND. We use this operation to combine sets of bounding boxes and their language elements, resulting in a new set with the merged attributes. In our dataset, we manually choose some meaningful attribute combinations and randomly generate many combinations for the objects. 

\noindent\textbf{Step 3: Prompt Generation.} After Step 2, we are able to determine the correspondence between combinations of language elements and a group of bounding boxes. However, getting valid natural language sentences to describe the objects can be expensive using human labor, and there is no guarantee of the desired variety. Large language model (LLM) have recently shown great potential in understanding logistics and producing sentences similar to those generated by humans. Therefore, we determine GPT3.5~\cite{gpt} as our language model. We prompt it with a request like ``Generate a sentence to describe the objects based on the following descriptions: \textit{pedestrians, moving, red, not in the left}'' where the italicized words represent the combination of elements. The LLM can respond with a meaningful description of the objects, such as ``The objects are red pedestrians, currently in motion, not situated on the left side." 
To guarantee accuracy, we will ask annotators to filter out any incorrect descriptions. 
We also prompt the LLM multiple times to generate multiple descriptions.

\subsection{Dataset Statistics}
\label{Sec:statistic}

Thanks to nuScenes~\cite{nuscenes2019}, our language prompts provide a number of comprehensive descriptions for the objects in the 3D, surrounding, and temporal space. Besides, they cover diverse environments comprising pedestrian streets, public roads, and highways in the cities of Boston and Singapore. Furthermore, they encompass different weather conditions (\eg, rain and sun), and illumination (\eg, day and night).
To offer deeper insights into NuPrompt, we next present more quantitative statistics.

\noindent\textbf{Language Prompt.}
We manually label 23,368 language elements and utilize them in the combination of 16,761 unified descriptions through LLM.
Among them, there are 12,001 unique language elements and 12,001 unique language combinations. 
In total, the NuPrompt has 40,147 language prompts.
On average, each video within the dataset contains 47 language prompts. 
We show the word cloud of the top 100 words in Fig.~\ref{fig:dataset_summary}.
From the left figure, we can observe that NuPrompt dataset has a large number of words that describe driving object appearances, like `black',  `white' and `red', and locations, like `front', `left' and `right'.
Besides, some motion words like `walking', `moving', and `crossing' are also common descriptions.

\begin{figure*}[t]
\centering
	\includegraphics[width=0.9\linewidth]{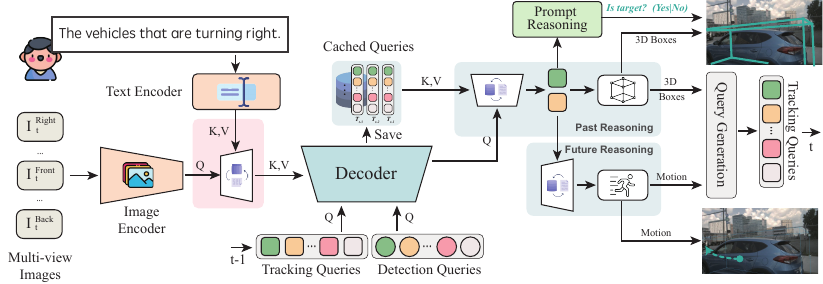}
	\caption{{Overall architecture of PromptTrack}. For each frame, the visual features and language prompt embedding are fused before being fed into the Transformer decoder.  Then past reasoning enhances and refines tacks by attending to cached historical queries,  and the future reasoning benefits cross-frame query propagation using predicted position. Lastly, the prompt reasoning branch predicts the prompt-referred tracks using binary classification. 
 }
	\label{fig:method}
\end{figure*}

\noindent\textbf{Referent Objects.}
In contrast to previous benchmarks that refer to 2D objects in modular images, another feature of NuPrompt is its surrounding 3D space.
This indicates that there are lots of objects crossing different views,  presenting improved simulation being closer to real driving scenes. 
More importantly, NuPrompt is designed to involve an arbitrary number of predicted objects.
The left of Figure~\ref{fig:dataset_summary} indicates that the majority of prompts describe between 1 and 10 instances, and can sometimes even exceed 20. 
 According to Table~\ref{table:dataset}, the average number of instances referred to by each prompt in our dataset is 7.4.
 In addition, the distribution of box is displayed on the right of Fig.~\ref{fig:dataset_summary}. 

\subsection{Benchmark Protocols}
\label{Sec:metrics}

\noindent\textbf{Task Definition.}
Together with NuPormpt, we formulate a straightforward yet challenging task that employs a language prompt as a meaningful signal to anticipate the pathways of objects. This task specifically includes multi-object tracking and movement prediction, based on prompts.

\noindent\textbf{Evaluation Metrics.}
To evaluate the similarity between the predicted tracklet and ground truth tracklet, we use Average Multiple Object Tracking Accuracy (AMOTA) as a primary metric~\cite{bernardin2008evaluating}.
However, unlike the original multi-object tracking task that averages AMOTA across different categories, the evaluation on NuPrompt is class-agnostic.
Hence, we calculate AMOTA for all prompt-video pairs and take the average of all these AMOTA values.
For a more detailed analysis, we also utilize Average Multi-Object Tracking Precision (AMOTP) and Identity Switches (IDS) metrics.
Additionally, we evaluate  trajectory prediction performance using  Average Displacement Error (ADE) and  Final
Displacement Error (FDE), following the works~\cite{pang2023standing,gu2023vip3d}.

\noindent\textbf{Data Split.}
The NuPrompt contains a total of 850 videos along with language prompts. Following nuScenes~\cite{nuscenes2019}, we split NuPrompt into training and validation set, which contain 700 videos and 150 videos, respectively.

\section{Method}
\label{sec:method}

Given multi-frame multi-view images and a natural language prompt, our goal is to track the described object. 
To accomplish this, we propose PromptTrack,  an end-to-end framework.  It modifies the query-based method PF-Track~\cite{pang2023standing} to adapt to the prompt input. 
Fig.~\ref{fig:method} shows the overall pipeline of PromptTrack. 
Note that PF-Track incorporates a past reasoning branch and a future reasoning branch which are based on the decoded queries. These branches aim to refine tack prediction using cached historical queries and improve cross-frame query propagation using motion localization prediction, respectively. In addition to these two branches, we propose a new prompt reasoning branch to predict the prompt-referred tracks.

\begin{table*}[t]
	\centering
	\small
	\resizebox{\textwidth}{!}{
		\begin{tabular}{l|c|ccccc|cc}
			\hline\thickhline
		Method & Decoder & AMOTA $\uparrow$ & AMOTP$\downarrow$ & RECALL$\uparrow$ & MOTA$\uparrow$ & IDS$\downarrow$ & ADE$\downarrow$ & FDE$\downarrow$  \\ 
		\hline
            \hline
            CenterPoint tracker& DETR3D & 0.079 & 1.820 & 19.6\% & 0.093 & 350 & - & -  \\
            CenterPoint tracker & PETR& 0.178 & 1.650& 29.1\% & 0.197 & 174 & - & -\\
            DQTrack & DETR3D & 0.186 & 1.641 & 30.7\% &  0.208 & 160 & 2.51 & 2.73\\
            DQTrack & Stereo &  0.198 & 1.625 & 30.9\% & 0.214 & 103 & 2.40 & 2.61\\
            DQTrack & PETRv2 & 0.234 & 1.545 & 33.2\% & 0.269 & 78 & 2.28 & 2.36  \\
            ADA-Track & PETR & 0.249 & 1.538 & 35.3\% &  0.270 & 67 & 2.20 & 2.31\\
            PromptTrack (Ours) & DETR3D & 0.202 & 1.615 & 31.0\% & 0.222 & 95 & 2.38 & 2.50  \\
            PromptTrack (Ours) & PETR &\textbf{0.259} &	\textbf{1.513} &	\textbf{36.6\%} & \textbf{0.280} & \textbf{26} & \textbf{2.17 }& \textbf{2.21} \\
	    \hline \thickhline
	\end{tabular} }
 	\caption{{Main results on NuPrompt}.  CenterPoint tracker~\cite{yin2021center}  is a heuristic-based tracking algorithm that utilizes different decoders such as DETR3D and PETR. 
  DQTrack, ADA-Track and our PromptTrack are end-to-end frameworks, which can also equip various decoders.
 $\uparrow$ and $\downarrow$ represent the direction of better performance about each metric. 
 }
	\label{table:sota}
\end{table*}

\subsection{Overall Architecture}
Formally, let $\bm{F}_t$ denote the extracted visual features at timestamp $t$, and $\bm{S}$ denote the encoded linguistic features.
To enrich the information on visual features, we first incorporate the visual features with the linguistic features in a multiplication way and form enhanced visual feature maps.
Specifically, we flatten two kinds of features and use cross-modal attention to encourage the feature fusion between $\bm{F}_t$ and $\bm{S}$, generating prompt-aware visual features $ \bm{F}_t'$:
\begin{equation}
\small
\begin{aligned}
    \bm{F}_t' = \textbf{CrossModalAttn}(&\text{Q}=\bm{F}_t, \text{K,V}=\bm{S}, \\
     &\text{PE}=\text{Pos}(\bm{S})),
     \label{eq: cross_modal}
\end{aligned}
\end{equation}
where $\text{Pos}$ means position embedding~\cite{vaswani2017attention}.

To capture different views and stereo information, we follow PETR~\cite{liu2022petr} to add 3D position embedding on $\bm{F}_t'$.
For notion simplicity, the position-augmented visual feature is also represented as $\bm{F}_t'$.
Then a set of 3D queries $\bm{Q}_t$ interact with $\bm{F}_t'$ via a stack of Transformer decoder layers, outputting updated queries $\bm{Q}_t^D$ and bounding boxes $\bm{B}_t^D$:
\begin{equation}
\small
    \bm{Q}_t^D, \bm{B}_t^D  = \textbf{Decoder} \left(\bm{Q}_t, \bm{F}_t'\right),
    \label{eq:decoder}
\end{equation}
where each input query $\bm{q}_t^i\!\in\! \bm{Q}_t$ means an object with a feature vector $\bm{f}_t^i$ and a 3D localization $\bm{c}_t^i$, \ie, $\bm{q}_t^i\!=\!\{\bm{f}_t^i, \bm{c}_t^i\}$.

To automatically link objects across different frames, the input queries $\bm{Q}_t$ merge track queries $\bm{Q}_t^{track}$ from the last frame. The box information from the last frame also provides an excellent spatial position prior to the current frame,  benefiting the model for accurately inferring the same object.
Besides, to capture new-born objects, a set of fixed 3D queries $\bm{Q}_t^{fixed}$, also called detection queries, are concatenated with track queries $\bm{Q}_t^{track}$ to generate $\bm{Q}_t$. 
Following the work~\cite{pang2023standing}, the number of fixed queries is set to 500.
As the first frame has no previous frames, we only utilize the fixed queries to detect objects.

\noindent\textbf{Past and Future Reasoning.}
After the Transformer Decoder, past and future reasoning are sequentially conducted for attending to historical embeddings and predicting future trajectory, respectively.
Formally, the past reasoning $\mathcal{F}^p$ integrates two decoded outputs $\bm{Q}_t^D$ and $\bm{B}_t^D$ as well as cached historical queries $\bm{Q}_{t-\tau_h:t-1}$ from past $\tau_h$ frames to produce refined queries $\bm{Q}_t^R$ and refined bounding boxes $\bm{B}_t^R$:
\begin{equation}
    \bm{Q}_t^R, \bm{B}_t^R  = \mathcal{F}^p \left(\bm{Q}_t^D, \bm{B}_t^D, \bm{Q}_{t-\tau_h:t-1} \right),
\end{equation}
where $\mathcal{F}^p$  has a cross-frame attention module for promoting history information integration across $\tau_h$ frames per object. 
Moreover, it also includes a cross-object attention module to encourage discriminative feature representation for each individual object.
The sequential cross-frame attention and cross-object attention modules lead to $\bm{Q}_t^R$.
A multi-layer perceptron (MLP) is used to predict coordinate residuals and adjust the object boxes, leading to $\bm{B}_t^R$.

Based on the refined results from past reasoning, the future reasoning $\mathcal{F}^f$ uses a cross-frame attention module to predict long-term trajectories $\bm{M}_{t: t+\tau_f}$ for next $\tau_f$ frames:
\begin{equation}
\small
    \bm{Q}_{t+1}^{track}, \bm{M}_{t: t+\tau_f}  = \mathcal{F}^f \left(\bm{Q}_t^R, \bm{Q}_{t-\tau_h:t-1} \right),
\end{equation}
where the position vectors of the refined queries $\bm{Q}_t^R$ is updated to generate $\bm{Q}_{t+1}^{track}$ according to the single-step movement $\bm{M}_{t: t+1}$. 
The main motivation of future reasoning is that as the ego-car goes forward, the reference position of all objects from the last frame has to be adjusted to align with the new ego-coordinates.
In summary, using past and future information can improve the quality of visible object tracking.
To further identify the object referred to in the prompt, we outline the new prompt reasoning as follows.

\subsection{Prompt Reasoning} 
Based on the past reasoning that tracks all visible objects, prompt reasoning focuses on grounding prompt-referred objects.
Since the refined queries $\bm{Q}_t^R$ from the past reasoning branch have integrated prompt embeddings, our prompt reasoning $\mathcal{F}^l$ ($l$ for `language')  can directly output the prompt-referred object probability $\bm{P}_t$:
\begin{equation}
\small
    \bm{P}_t  = \mathcal{F}^l \left(\bm{Q}_t^R \right),
\end{equation}
where $\mathcal{F}^l$ is an MLP with two fully-connected layers.  $\bm{P}_t$ is a binary probability that indicates whether the output embedding is a prompt-referred object.



\begin{table}[t]
\label{table:ablation}
    \begin{center}
    \small
     \begin{subtable}[t]{0.49\textwidth}
        \centering
		\begin{tabular}{c|ccc}
			\hline\thickhline
			Prompt Fusion & AMOTA $\uparrow$ & AMOTP$\downarrow$ & RECALL$\uparrow$    \\ 
			\hline
            \hline
            \textit{w/o}  Prompt Fusion & 0.073 & \textbf{1.398} & \textbf{43.5\%}   \\
            Two-way Fusion & 0.249 &1. 530& 36.7\%   \\
            Prompt as Query & 0.054 & 1.723 & 21.6\% \\
              \rowcolor[gray]{0.9}
              Ours &\textbf{0.259}&{1.513} & {36.6\%}  \\
			\hline \thickhline
	\end{tabular}
    \end{subtable}
\caption{
Different prompt-visual fusion methods. 
 \textit{w/o} Prompt Fusion' represents removing prompt reasoning and using all tracking instances as prompt prediction. `Two-way Fusion' and `Prompt as Query' are fusion variants. 
 }
 \end{center}
\end{table}

\begin{table}[t]
\label{table:ablation_past}
    \begin{center}
    \small
     \begin{subtable}[t]{0.49\textwidth}
        \centering
        \begin{tabular}{c|cccc}
			\hline\thickhline
			Past \& Future& AMOTA $\uparrow$ & AMOTP$\downarrow$ & RECALL$\uparrow$  \\ 
			\hline
            \hline
               \textit{w/o} Past Reason & 0.239 &1.543 & 33.1\%\\
               \textit{w/o} Future Reason & 0.247 &1.523 & 34.9\% \\
              \rowcolor[gray]{0.9}
              Ours & \textbf{0.259}& 1.513 & 36.6\% \\
			\hline \thickhline
	\end{tabular}
    \end{subtable}
\caption{
Ablation study on past and future reasoning.
 }
 \end{center}
\end{table}

\subsection{Instance Matching and Loss}

Our method views the 3D detection as a set prediction problem following query-based methods~\cite{carion2020end,wu2023onlinerefer}, so it requires one-to-one matching before calculating loss. The tracking queries $\bm{Q}_t^{track}$ and their ground-truth have been matched when propagating queries. As for the fixed queries $\bm{Q}_t^{fixed}$, we match them with new-born objects using a bipartite graph matching, following MOTR~\cite{zeng2022motr}.
During the process of matching, we only use the queries $\bm{Q}_t^D$ from decoder outputs, and then implement the correspondence in the overall loss:
\begin{equation}
\begin{aligned}
        \mathcal{L} = \lambda_{cls}^D\mathcal{L}_{cls}^D + \lambda_{box}^D\mathcal{L}_{box}^D + +  \lambda_{f}\mathcal{L}_{f}  \\
        +  \lambda_{cls}^R\mathcal{L}_{cls}^R + \lambda_{box}^R\mathcal{L}_{box}^R  + \lambda_{p}\mathcal{L}_{p},
\end{aligned}
\end{equation}
where $\mathcal{L}_{cls}$s are Focal loss~\cite{lin2017focal} for classification, $\mathcal{L}_{box}$s are L1 loss for bounding box regression, $\mathcal{L}_{f}$ is L1 loss for motion prediction. 
The settings of their weights $\lambda$s follow PF-Track ~\cite {pang2023standing}.
Besides, our prompt reasoning loss $\mathcal{L}_{p}$ is also  Focal loss, weighted by $\lambda_p$.


\section{Experiment}

In this section, we conduct a set of experiments on our proposed benchmark NuPrompt. 
For implementation details, we follow the settings of PF-Track~\cite{pang2023standing}.

\subsection{Main Results}
Since there are no existing methods for the new prompt-based prediction task, we modify the existing \textbf{open-sourced} camera-only tracking models for state-of-the-art comparison.
First, most previous camera-only tracking methods follow a heuristic-based pattern~\cite{liu2023bevfusion,wang2023exploring},  which include two parts: basic detectors and trackers.
Following them, we utilize two state-of-the-art  single-frame object detectors, \ie, DETR3D~\cite{wang2022detr3d} and PETR~\cite{liu2022petr}, and a popular tracker, CenterPoint tracker~\cite{yin2021center}. 
The powerful tracker has been used in many state-of-the-art detection methods from the leaderboard of nuScenes tracking task~\cite{liu2023bevfusion,wang2023exploring}.
To distinguish the prompt-referred objects, the basic detectors incorporate the implementation of our prompt fusion module.
Second, we incorporate two end-to-end camera-based 3D multi-object tracking models DQTrack~\cite{li2023end} and ADA-Track~\cite{ding2024ada}, which  utilizes diverse decoders like DETR3D~\cite{wang2022detr3d}, Stereo~\cite{li2023end},  PETRv2~\cite{liu2023petrv2}, or PETR~\cite{liu2022petr}.
This approach stands apart from the previously mentioned heuristic-based methods that rely on a two-step inference process.
We seamlessly integrate our prompt reasoning module within DQTrack to enable the prediction of objects specifically referred to in prompts.
For all other configurations, we adhere to the official settings and guidelines.
Third, we evaluate our PromptTrack framework by substituting its fundamental detector with DETR3D.
On top of NuPrompt, we test the proposed PromptTrack and these competitors, and the main results are shown in Table~\ref{table:sota}. 
As seen, PromptTrack achieves 0.259 on AMOTA, outperforming other counterparts across the majority of metrics.

In addition to prompt-refereed tracking evaluation, we assess the performance of motion prediction by forecasting over eight frames.  
As object motion cannot be effectively tested by single-frame object detectors, PromptTrack is the only model we examine. 
From  Table~\ref{table:sota}, PromptTrack with PETR detector results in an ADE of 2.17 and a FDE of 2.21, showing impressive performance. 
For inference speed, we test PromptTrack using VOV backbone on one Nividia A100 GPU over the validation set. PromptTrack achieves 7.7 FPS.

 \begin{table}[t!]
 \small
    \begin{center}
    \begin{subtable}[t]{0.48\textwidth}
        \centering
		\begin{tabular}{c|cccc}
			\hline\thickhline
			Prompt Type & AMOTA $\uparrow$ & AMOTP$\downarrow$ & RECALL$\uparrow$    \\ 
			\hline
            \hline
              Language Elements & 0.271 & 1.498 & 38.9\%  \\
              Unified Descriptions & 0.242 &1.523 & 35.7\%   \\
              \rowcolor[gray]{0.9}
              Total Prompts & {0.259}& 1.513 & 36.6\%  \\
			\hline \thickhline
	\end{tabular} 
        \label{table:3a}
    \end{subtable}
\end{center}
        \caption{
{Ablation studies}  on different prompt types. 
    }
\end{table}

  \begin{figure}[t]
	\centering
	\resizebox{0.48\textwidth}{!}
	{
		\includegraphics[width = 16cm]{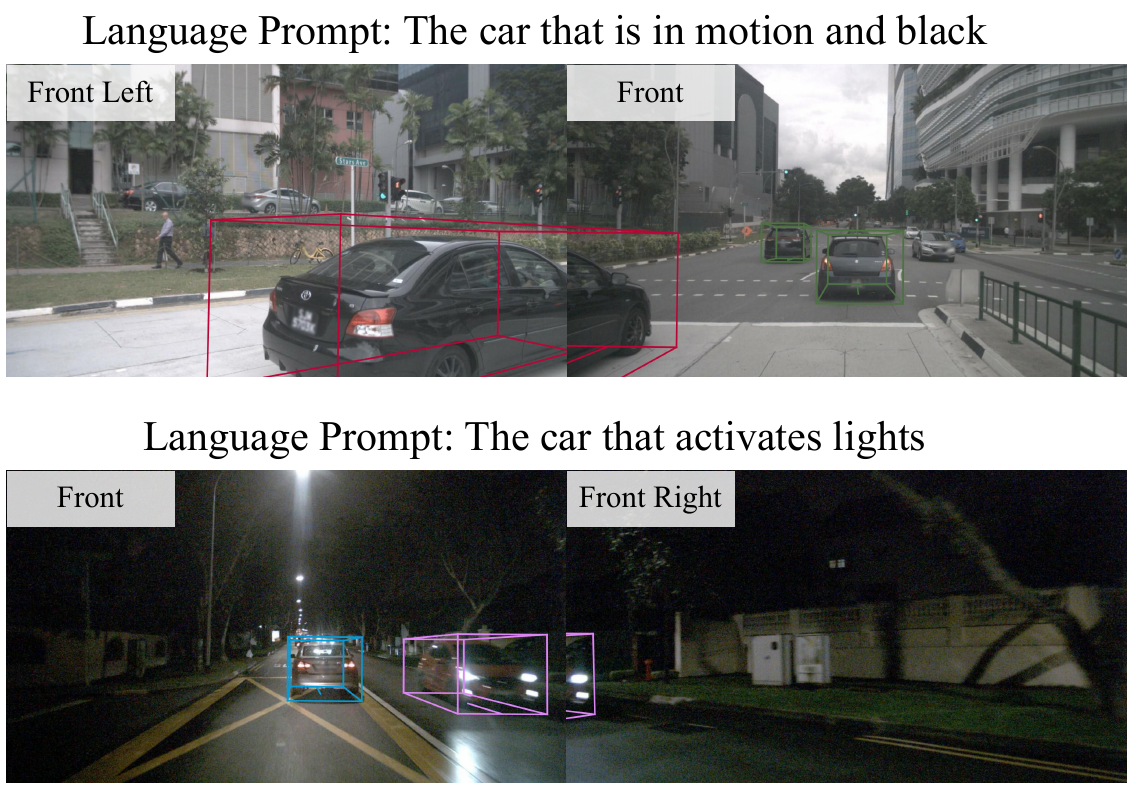}
	}
	\caption{{Qualitative results} of PromptTrack on NuPrompt.
	}
	\label{fig:results}
\end{figure} 

\subsection{Ablation Studies}
\label{sec:ablation}

\noindent\textbf{Prompt Fusion Methods.} 
Table 3 showcases various prompt-visual fusion strategies. The first is to eliminate the prompt fusion module and use all predicted tracking instances as our prompt prediction (\textit{w/o} Prompt Fusion).
Second, we can improve upon our current approach of utilizing language prompt features solely to augment visual features (one-way fusion) by introducing a two-way fusion method. Specifically, before enhancing the visual features, we generate visual-enhanced language features, which bring a bidirectional interaction between the two modalities.
Third, we evaluate another variant of prompt-visual fusion, only adding prompt embedding with the decoder queries (represented as `Prompt as Query').
These variants, though implemented, do not yield notable performance gains.
In summary, our prompt-visual fusion strategy proves to be highly effective for the given task.

\noindent\textbf{Past \& Future Reasoning.}
In Table 4, it is evident that neglecting either past or future reasoning leads to diminished performance, emphasizing the importance of incorporating both reasoning branches.

 \begin{table}[t!]
 \small
    \begin{center}
    \begin{subtable}[t]{0.48\textwidth}
        \centering
		\begin{tabular}{p{26mm}|ccc}
			\hline\thickhline
			Prompt Number& AMOTA $\uparrow$ & AMOTP$\downarrow$ & RECALL$\uparrow$    \\ 
			\hline
            \hline
              20\% & 0.210 &1.471 & 31.3\%  \\
              50\% & 0.247 &1.511 & 36.1\%  \\
              \rowcolor[gray]{0.9}
              100\% &{0.259}&{1.513} & {36.6\%}\\
			\hline \thickhline
	\end{tabular} 
        \label{table:3b}
    \end{subtable}
\end{center}
        \caption{
{Ablation studies}  on different prompt numbers.
    }
\end{table}

\noindent\textbf{Prompt Type \& Number.}
We test our PromptTrack on language elements and unified descriptions (see details in \S\ref{Sec:collection}), as presented in Table 5.
Notably, the model performance on the unified descriptions exhibits a decline compared to that on the language elements, suggesting that the integration of the combined prompts poses certain challenges.
We additionally experiment to examine the impact of varying prompt numbers by randomly sampling 20\%, 50\%, and 100\% from the training set, as shown in Table 6. 
A clear trend emerges: as the number of prompts increases, there is a corresponding enhancement in model performance, highlighting the positive correlation between prompt quantity and model efficacy.

\subsection{Qualitative Results}
We visualize two typical qualitative results in Fig.~\ref{fig:results}.
As seen, our PromptTrack can detect and track prompt-referred targets accurately under various challenging situations, like crossing different views and varying object numbers.


\section{Conclusion}

In this work, we presented NuPrompt, the first large-scale language prompt set designed specifically for 3D perception in autonomous driving.
NuPrompt provides numerous precise and fine-grained 3D object-text pair annotations.
As a result, we designed a simple yet challenging prompt-driven object trajectory prediction task, \ie, tracking objects and forecasting their motion using a language prompt as a semantic cue.
To solve this problem, we further proposed an end-to-end prompt-based tracking model with prompt reasoning modification on PF-Track, called PromptTrack.
After conducting a set of experiments on NuPrompt, we verified the effectiveness of our algorithm.

\section*{Acknowledgments}
This work was supported by the FDCT grants 0102/2023/RIA2, 0154/2022/A3, and 001/2024/SKL, the SRG2022-00023-IOTSC grant and the MYRG-CRG2022-00013-IOTSC-ICI grant.

\bibliography{aaai25}

\end{document}